\documentclass[letterpaper,10pt,conference]{ieeeconf}

\IEEEoverridecommandlockouts

\usepackage{amsmath,amssymb}
\usepackage{graphicx}
\usepackage{xcolor}
\usepackage{url}
\usepackage[hidelinks]{hyperref}

\newcommand{\method}{GAUGE}
\newcommand{\vect}[1]{\boldsymbol{#1}}
\newcommand{\RealOffsetRawRmse}{0.05288}
\newcommand{\RealOffsetGaugeRmse}{0.02678}
\newcommand{\RealOffsetReduction}{49.4\%}
\newcommand{\RealArcRawRmse}{0.05523}
\newcommand{\RealArcGaugeRmse}{0.02298}
\newcommand{\RealArcReduction}{58.4\%}

\newcommand{\RealCalibrationProbeCount}{16}

\newcommand{\RealNavigationSuccess}{16/16}

\title{\LARGE \bf GAUGE: Planner-Conditioned Active Calibration of Opaque Quadruped Velocity Interfaces}

\author{Tianhao Zang$^{1,2}$, Zihan Liu$^{3}$, Shanze Wang$^{4}$, Liyou Luo$^{5}$,\\
Xingjian Xie$^{3}$, Chengtai Li$^{6}$, Wei Zhang$^{1,*}$
\thanks{$^{1}$College of Information Science and Technology, Eastern Institute of Technology, Ningbo, P. R. China.}
\thanks{$^{2}$University of Maryland, USA.}
\thanks{$^{3}$School of Computer Science, University of Nottingham, UK.}
\thanks{$^{4}$The Hong Kong Polytechnic University, Hong Kong.}
\thanks{$^{5}$Xi'an Jiaotong-Liverpool University, China.}
\thanks{$^{6}$Eastern Institute of Technology, Ningbo, P. R. China.}
\thanks{$^{*}$Corresponding author. Email: \texttt{zhw@eitech.edu.cn}.}}

\begin{document}
\maketitle
\thispagestyle{empty}
\pagestyle{empty}

\begin{abstract}
In this paper, we present a Goal-Aware Uncertainty-Guided Exploration (GAUGE)
framework for planner-conditioned active calibration of opaque quadruped
velocity interfaces. Commercial quadrupeds commonly expose planar-velocity
commands, but the underlying locomotion controller remains inaccessible and
can produce systematic discrepancies between commanded and realized motion.
A navigation planner typically uses a structured subset of the command envelope.
GAUGE maintains a Bayesian command-to-motion model
and selects authorized trials according to their expected reduction of
posterior epistemic uncertainty under the planner-induced command distribution.
The resulting posterior supports validation-based stopping, bounded inverse
compensation, and task-relevant recalibration after detected interface shifts.
In three controlled response families, GAUGE reaches the joint criterion for
task-facing accuracy and uncertainty with fewer trials than passive, D-optimal,
and task-agnostic alternatives. Across six held-out Isaac Sim navigation maps,
it meets the declared noninferiority margins against dense calibration.
Code is available at \url{https://github.com/EurekaZang/CalibAgent}.
\end{abstract}

\section{Introduction}

High-level body-velocity commands provide a common interface between navigation
planners and quadruped locomotion controllers
\cite{hwangbo2019agile,taouil2023blackbox}. This separation allows a planner to
request forward, lateral, and yaw motion without controlling individual joints
or gait phases. On commercial platforms, however, the controller that converts
these requests into body motion is often proprietary and cannot be inspected or
modified by the user. The exposed command is therefore convenient, but it is
not necessarily a calibrated motion variable. Gain error, dead zones,
saturation, cross-axis coupling, payload, terrain, and controller settings can
all change the motion realized from the same nominal request. A quadruped may
remain stable while executing a systematically different velocity from the one
assumed by its navigation planner. We refer to this discrepancy as the
command--motion gap. Because the planner predicts tracking and obstacle
clearance through the exposed interface, the gap directly perturbs the control
variable used for navigation. Calibrating that interface is consequently
important whenever a mature locomotion controller must be deployed without
replacing or opening its internal control stack.

Existing methods address related deployment mismatch by modifying or estimating
quantities inside the locomotion system. Simulation-trained policies improve
transfer through actuation modeling, randomized training, curricula, and
proprioceptive feedback
\cite{hwangbo2019agile,lee2020terrain,rudin2022minutes,curtis2025flow}.
Adaptation policies infer latent environment or embodiment states from recent
interaction histories \cite{kumar2021rma,fey2024adapt,li2026embodiment}, while
online system identification and Bayesian optimization update model parameters
or controller gains \cite{sun2021unknown,widmer2023safe,haack2025adaptive}.
These approaches are effective when the policy, controller, or model is
available as an adaptation target. An opaque commercial controller presents a
different boundary: only commands and realized motion are visible. External
response models address this boundary by identifying an input--output map from
measured motion \cite{li2022vio,taouil2023blackbox}, but collecting uniform or
exhaustive coverage over all command combinations is expensive on hardware.
Classical active calibration reduces this burden by selecting trials according
to parameter observability or information criteria
\cite{hollerbach1996calibration,calafiore2001dynamic,sun2008observability,sun2008active}.
Task- and goal-oriented design instead evaluates measurements through uncertainty
in a downstream quantity \cite{carrillo2013task,attia2018goal}, and active
legged-system identification ranks excitation by information about physical
simulation parameters \cite{sobanbabu2025spiactive}. These lines leave a
specific interface-calibration problem: when the controller and physical model
are both unavailable, how should a small number of executable commands be
chosen when their value is determined by the velocity distribution used by the
downstream planner?

In this work, we approach opaque-interface calibration from the planner side.
The central insight is that calibration should minimize uncertainty over the
commands the planner is expected to use, rather than distribute accuracy
uniformly over every command the robot can execute. The planner induces a
deployment measure from nominal waypoint-following commands, and this measure
defines task-facing uncertainty in the external command-to-motion response.
Each calibration trial is then valued by its expected reduction of that
uncertainty. This changes the objective of data acquisition without requiring
access to the hidden dynamics or controller.

\begin{figure*}[t]
  \centering
  \includegraphics[width=\textwidth]{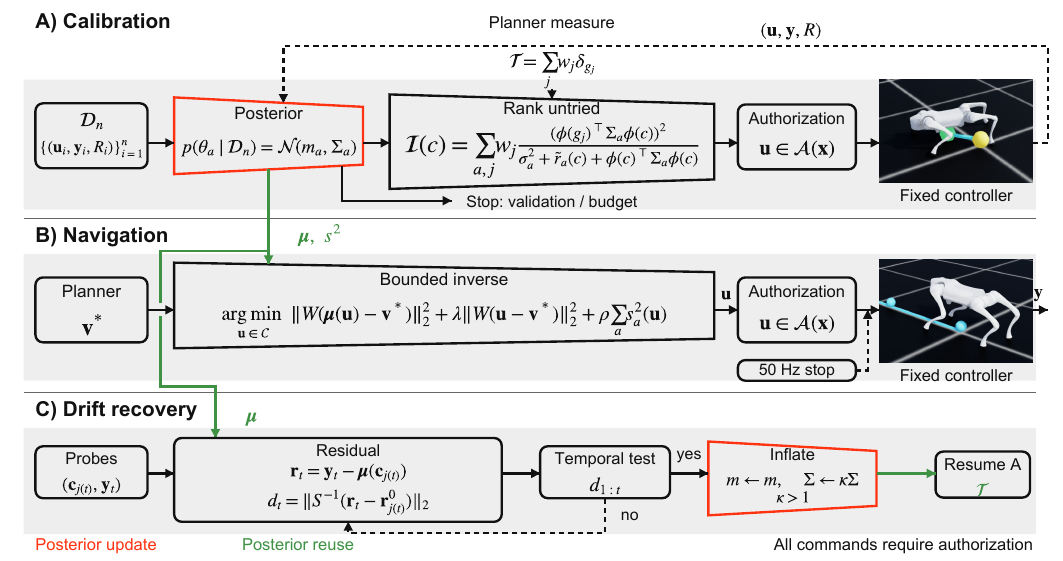}
  \caption{GAUGE framework. (A) IVR, weighted by the planner-induced $\mathcal{T}$, ranks untried commands in the finite pool $\mathcal{C}$; authorization selects the highest-ranked feasible command. Valid observations update the posterior. (B) The posterior supports bounded inversion. (C) Frozen $\mu$, $r_j^0$, and $S$ support temporal shift detection. Recovery retains $m$, inflates $\Sigma$, and resumes (A) with the same $\mathcal{T}$. Validation or hard budgets end calibration and recovery. All commands require authorization; the controller is fixed. Images are simulator captures.}
  \label{fig:pipeline}
\end{figure*}

\method\ (Goal-Aware Uncertainty-Guided Exploration) operationalizes this
principle as a calibration and recalibration pipeline
(Fig.~\ref{fig:pipeline}). It begins with a small set of measured
command--motion pairs and fits a coupled Bayesian response model that represents
gain error and off-axis response while tracking posterior epistemic
uncertainty. Task-weighted integrated variance reduction ranks the next
candidate command under the planner-induced measure. A separate hard filter
remains the sole authority for executing that command, keeping statistical
information value distinct from operational feasibility. After each accepted
trial, the posterior is updated and evaluated on fixed validation commands.
Calibration stops when the response error and task-facing uncertainty satisfy
their declared criteria or when the interaction budget is exhausted. During
deployment, bounded inverse compensation maps a desired body motion back to an
executable command using the calibrated response. A residual-monitoring
extension detects when the frozen response has changed, inflates posterior
uncertainty, and reopens the same planner-conditioned acquisition process. The
same response posterior therefore connects initial calibration, command
compensation, and post-shift recovery without modifying the low-level
locomotion controller.

We evaluate this principle through physical response measurements, controlled
acquisition studies, shift experiments, and held-out navigation. On 183 passive
Unitree Go2 trials, a coupled-affine response model reduces leave-one-session-out
RMSE from 0.066 for the identity command map to 0.030. Online calibration in two independently fitted route deployments reduces
held-out prediction RMSE by 49.4\% and 58.4\%, with lower pointwise error on all
16 matched validation commands. In three controlled response families, task
IVR reaches the first crossing of the joint task-facing accuracy--uncertainty
criterion in 18.7 trials, compared with 23.0 for D-optimal design and 25.7 for
task-agnostic IVR. At a fixed 24-trial budget, expanding evaluation to broad
uniform support reverses the held-out-error ordering, confirming that the gain
comes from matching the acquisition objective to the planner distribution.
Controlled simulation and a scheduled, known Go2 gain-and-coupling perturbation
further show the intended early-recovery trade-off: planner conditioning lowers
early task-facing error, whereas passive updating can attain lower terminal
error after the limited-budget window. Finally, a 12-trial \method\ calibration
satisfies the declared success and capped-time noninferiority margins against a
30-trial dense reference across six held-out Isaac Sim navigation maps.
Together, these experiments test the calibration principle at the
response-model, acquisition, recovery, and navigation levels.

\noindent\textbf{Contributions:}
\begin{itemize}
  \item \textbf{Planner-conditioned calibration objective.}
  We formulate external-response calibration under a planner-induced command
  measure. Expected reductions in task-weighted posterior epistemic uncertainty
  determine the value of each trial.
  \item \textbf{Operational calibration lifecycle.}
  We combine active acquisition with independent execution authorization,
  validation-based stopping, and bounded inverse compensation. The same
  task-weighted objective supports reacquisition after interface shifts.
  \item \textbf{Source attribution and support boundary.}
  Task-agnostic ablations and a support-mismatch control identify why and where
  planner conditioning saves trials. Go2 measurements and held-out Isaac
  navigation link calibration to physical and downstream behavior.
\end{itemize}

\section{Related Work}

\subsection{Estimation Targets}

Legged adaptation and system identification differ first in what they make
explicit. Model-based and data-driven schemes estimate physical quantities,
unknown dynamics, or local closed-loop response models and feed those estimates
back into control \cite{sun2021unknown,fey2024adapt,haack2025adaptive}.
Controller-level approaches optimize feedback gains \cite{widmer2023safe},
whereas adaptive policies infer latent environment or embodiment states from
recent interaction histories \cite{kumar2021rma,li2026embodiment}. Other sim-to-real pipelines
place robustness in the learned policy through actuation modeling, randomized
training, curricula, or proprioceptive feedback, without exposing a physical
parameter estimate at deployment
\cite{hwangbo2019agile,lee2020terrain,rudin2022minutes}. Across these approaches,
the modified or inferred quantities remain internal to the model, controller,
or policy.

External-interface identification adopts an input--output estimand.
Velocity-control kinematics can be calibrated jointly with odometry from
commanded and measured motion \cite{li2022vio}; black-box quadruped response
models similarly map commands to center-of-mass and footstep behavior for
planning \cite{taouil2023blackbox}. In this view, the hidden controller remains
fixed, and the observable pair $(\vect{u},\vect{y})$, rather than a named
internal parameter, defines the object of calibration.

\subsection{Trial-Selection Objectives}

Once the estimand is fixed, experimental design determines which reduction in
uncertainty is valuable. Classical robot calibration uses calibration indices,
optimal excitation trajectories, observability, and alphabetic parameter
criteria
\cite{hollerbach1996calibration,calafiore2001dynamic,sun2008observability,sun2008active}.
Information-theoretic placement likewise maximizes entropy or mutual
information over a modeled field \cite{krause2008sensor}, while exhaustive
command enumeration supplies global coverage for an opaque response map
\cite{taouil2023blackbox}. These objectives prioritize parameter recovery or
coverage of the modeled domain.

Task- and goal-oriented designs instead evaluate a trial through a downstream
quantity. Manipulator calibration can select configurations by uncertainty at
the tool center point \cite{carrillo2013task}, and goal-oriented Bayesian design
optimizes posterior uncertainty in a designated quantity of interest
\cite{attia2018goal}. In legged system identification, active excitation can be
ranked by trajectory Fisher information about physical simulation parameters
\cite{sobanbabu2025spiactive}. The literature therefore separates two
decisions: what is estimated and over which measure its uncertainty is valued.
For navigation-facing interface calibration, that downstream measure is the
command distribution induced by the planner.

\section{GAUGE: Planner-Conditioned Interface Calibration}

\subsection{Problem Setting and Planner-Induced Task Measure}

The low-level controller is fixed and opaque. A trial applies a constant
body-frame command $\vect{u}\in\mathcal{U}\subset\mathbb{R}^{3}$ and extracts a
robust steady-motion observation $\vect{y}\in\mathbb{R}^{3}$ with measurement
covariance $R$. Invalid observations are rejected before the dataset
$\mathcal{D}_n=\{(\vect{u}_i,\vect{y}_i,R_i)\}_{i=1}^{n}$ is updated. A finite
design pool $\mathcal{C}$ defines the modeled command envelope. Runtime state
$\vect{x}$ induces a separate hard-authorized subset
$\mathcal{A}(\vect{x})\subseteq\mathcal{C}$. The authorized subset governs
feasibility, while the task measure below assigns information value.

The planner induces the discrete deployment command measure
$\mathcal{T}=\sum_{j=1}^{G}w_j\delta_{\vect{g}_j}$, where
$\vect{g}_j\in\mathcal{U}$ is a support command obtained by discretizing
nominal waypoint-planner outputs, $\delta_{\vect{g}_j}$ is a unit point mass,
and $w_j\geq0$ is the deployment probability assigned to that command, with
$\sum_{j=1}^{G}w_j=1$. The weights may be estimated from nominal planner-output
frequencies or specified by the declared route mixture. Let
$\vect{\phi}(\vect{u})\in\mathbb{R}^{p}$ be the response-model feature vector and
let
$\vect{\Sigma}_{a,n}:=\operatorname{Cov}(\vect{\theta}_a\mid\mathcal{D}_n)
\in\mathbb{R}^{p\times p}$ be the posterior covariance of the coefficient
vector for output axis $a\in\{1,2,3\}$ after $n$ trials. The task-facing
epistemic uncertainty is
\begin{equation}
 \mathcal{V}_{\mathcal{T}}(\mathcal{D}_n)=
 \sum_{a=1}^{3}\sum_{j=1}^{G} w_j
 \vect{\phi}(\vect{g}_j)^\top\vect{\Sigma}_{a,n}
 \vect{\phi}(\vect{g}_j).
 \label{eq:taskvariance}
\end{equation}
Calibration aims to reduce this deployment-facing quantity under the available
trial budget. For a fixed budget $B$, the acquisition policy selects authorized
commands to minimize the expected value of
$\mathcal{V}_{\mathcal{T}}(\mathcal{D}_B)$. Uniform accuracy over
$\mathcal{C}$ is a different calibration objective, aligned with deployment
objectives whose support spans the authorized envelope.

\subsection{Bayesian External-Response Model}

\method\ represents the exposed command--motion interface with a coupled
Bayesian response model. Its posterior covariance evaluates the task-facing
objective in Eq.~\eqref{eq:taskvariance} and supplies the acquisition rule below.
For command $\vect{u}$ and measured steady velocity $\vect{y}\in\mathbb{R}^3$,
the coupled affine basis is $[1,v_x,v_y,\omega_z]$. A diagonal control zeros
off-axis coefficients. The nonlinear 13-term basis adds pairwise products and
$[v_k-\tau_k]_+,[-v_k-\tau_k]_+$, with
$\vect{\tau}=[0.15,0.10,0.25]$. Non-intercept columns are standardized once on
the design pool, and the closed-loop prior regularizes the linear projection
toward identity.

For observation $i$ and output axis $a$, the model is
\begin{equation}
 y_{i,a}=\vect{\theta}_a^\top\vect{\phi}(\vect{u}_i)+\epsilon_{i,a},
 \quad
 \epsilon_{i,a}\sim\mathcal{N}(0,\sigma_a^2+r_{i,a}),
 \label{eq:model}
\end{equation}
where $r_{i,a}=(R_i)_{aa}$ is the measurement-pipeline variance and $\sigma_a^2$
is the frozen process variance. Given prior
$\mathcal{N}(\vect{m}_{a,0},\vect{\Sigma}_{a,0})$, define
$\vect{\Lambda}_{a,0}=\vect{\Sigma}_{a,0}^{-1}$ and
$\vect{\eta}_{a,0}=\vect{\Lambda}_{a,0}\vect{m}_{a,0}$. Sequential updates use
\begin{align}
 \vect{\Lambda}_{a,n} &= \vect{\Lambda}_{a,n-1}+
 \frac{\vect{\phi}(\vect{u}_n)\vect{\phi}(\vect{u}_n)^\top}
 {\sigma_a^2+r_{n,a}}, \\
 \vect{\eta}_{a,n} &= \vect{\eta}_{a,n-1}+
 \frac{\vect{\phi}(\vect{u}_n)y_{n,a}}{\sigma_a^2+r_{n,a}},
 \label{eq:update} \\
 \vect{\Sigma}_{a,n} &= \vect{\Lambda}_{a,n}^{-1},\qquad
 \vect{m}_{a,n}=\vect{\Sigma}_{a,n}\vect{\eta}_{a,n} . \notag
\end{align}
The predictive mean is
$\mu_{a,n}(\vect{u})=\vect{m}_{a,n}^\top\vect{\phi}(\vect{u})$. The response variance
used by bounded inversion is
$s_{a,n}^2(\vect{u})=\vect{\phi}(\vect{u})^\top\vect{\Sigma}_{a,n}
\vect{\phi}(\vect{u})+\sigma_a^2$. Coverage of a future measured response also
adds the corresponding observation variance. Acquisition uses only the reducible
epistemic term. Coupled input features represent cross-axis response while the
three output posteriors remain conditionally independent. We omit the posterior
index $n$ in deployment expressions when the active dataset is unambiguous.

\subsection{Planner-Conditioned Trial Selection}

The acquisition rule of \method\ values each trial by its expected reduction
of uncertainty under the planner-induced command distribution. For candidate
$\vect{c}\in\mathcal{C}$, task-weighted integrated variance reduction (IVR),
obtained by a Sherman--Morrison update, gives the conditional one-step reduction
in Eq.~\eqref{eq:taskvariance}:
\begin{equation}
 \mathcal{I}(\vect{c})=
 \sum_{a=1}^{3}\sum_{j=1}^{G} w_j
 \frac{\left(\vect{\phi}(\vect{g}_j)^\top\vect{\Sigma}_{a,n}
 \vect{\phi}(\vect{c})\right)^2}
 {\sigma_a^2+\tilde r_a(\vect{c})+
 \vect{\phi}(\vect{c})^\top\vect{\Sigma}_{a,n}
 \vect{\phi}(\vect{c})}.
 \label{eq:ivr}
\end{equation}
The optional $\tilde r_a(\vect{c})$ predicts the observation-noise variance of
an unexecuted candidate. After execution, the posterior update uses the measured
$r_{n+1,a}$. The selector ranks untried candidates by
$\mathcal{I}(\vect{c})$, and greedy batches use outcome-independent covariance
updates. Hard authorization removes infeasible candidates before the
highest-ranked remaining command is executed.

\subsection{Deploying the Calibrated Interface}

Three operations connect the task-conditioned posterior to deployment:
authorization, validation stopping, and bounded inversion. The hard filter
checks state validity, localization, battery, attitude, base
height, command and slew bounds, combined normalized linear--angular load, and
workspace occupancy. It
fails closed if no ranked candidate remains, and an independent 50~Hz monitor
can zero commands. Statistical value cannot relax an execution limit.

Stopping requires minimum coverage and three consecutive checks of held-out
validation RMSE and task-facing uncertainty. Hard budgets take priority. During
deployment, the inverse searches the bounded candidate set for planner target
$\vect{v}^{*}$ and minimizes
\begin{equation}
 \|W(\vect{\mu}(\vect{u})-\vect{v}^{*})\|_2^2+\lambda
 \|W(\vect{u}-\vect{v}^{*})\|_2^2+\rho\sum_a s_a^2(\vect{u}),
 \label{eq:inverse}
\end{equation}
where $W$ normalizes the three axes and $\lambda,\rho\geq0$ trade command
regularization against predictive risk. The selected command then passes
through the hard filter.

\subsubsection*{Deployment Extension}
Authorized probe commands can monitor whether the calibrated interface has
shifted. Commissioning freezes residual signatures
$\vect{r}^{0}_{j}=\vect{y}_{j}-\vect{\mu}(\vect{c}_{j})$ and a diagonal scale
matrix $S$. A later probe produces
$\vect{r}_t=\vect{y}_t-\vect{\mu}(\vect{c}_{j(t)})$ and standardized deviation
$d_t=\|S^{-1}(\vect{r}_t-\vect{r}^{0}_{j(t)})\|_2$. When a
deployment-specific temporal test latches a shift, the posterior mean is
retained, its covariance is inflated, and the planner-conditioned acquisition
rule in Eq.~\eqref{eq:ivr} is reopened under the same task measure. Validation
or a hard recovery budget closes the extension. Section~IV-C specifies the
experimental detector and recovery settings.

\begin{figure*}[t]
  \centering
  \includegraphics[width=\textwidth]{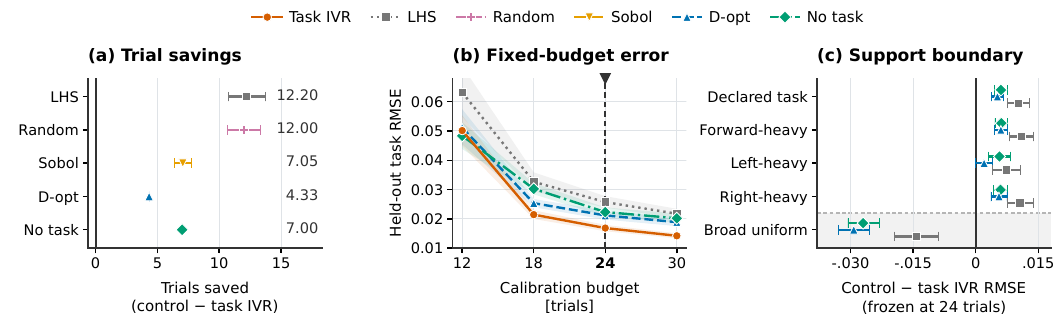}
  \caption{Core evidence for planner-conditioned acquisition. The independent
  unit is one of 20 paired seeds; the affine, dead-zone, and heteroscedastic
  response families are averaged within each seed. (a) Control-minus-task-IVR
  first-crossing trials for the joint RMSE $\leq .04$ and epistemic-variance
  $\leq .0015$ criterion; task IVR crosses at 18.7 trials on average. (b)
  Held-out task RMSE across fixed budgets, with the 24-trial acquisitions marked
  for the support test. (c) The same frozen 24-trial acquisitions re-evaluated
  under alternative command supports; positive values favor task IVR, while
  broad-uniform support reverses the ordering. Bands and error bars are paired
  95\% bootstrap intervals over seeds.}
  \label{fig:calibration}
\end{figure*}

\section{Experiments and Results}

The evaluation tests one central hypothesis: under a structured planner-induced
command distribution, task-conditioned acquisition reaches task-facing response
accuracy with fewer interactions than passive, parameter-oriented, or
task-agnostic designs. Response-model comparisons establish the calibration
substrate. Task-agnostic and
support-expansion controls isolate the effect and boundary of task weighting.
Shift experiments test reuse of the same objective, and navigation measures its
downstream consequence. Hardware experiments establish physical response-model
behavior, while paired synthetic and fixed-configuration Isaac experiments
provide replicated comparisons of acquisition and recovery mechanisms.

\subsection{Coupled Modeling Improves Go2 Response Prediction}

The passive Go2 dataset contains 183 valid trials from three sessions,
localized with a Livox MID-360 running FAST-LIO~\cite{xu2021fastlio}. In
leave-one-session-out evaluation, modeling cross-axis terms is more important
than adding an intercept. The coupled-affine model reduces pooled
root-mean-square error (RMSE) from 0.066 for the identity map to 0.030 and
outperforms the diagonal-affine model by 34.1\%.
The fitted yaw-to-lateral coefficient remains stable across held-out sessions,
indicating systematic rather than session-specific coupling.

\begin{figure*}[t]
  \centering
  \includegraphics[width=0.98\textwidth]{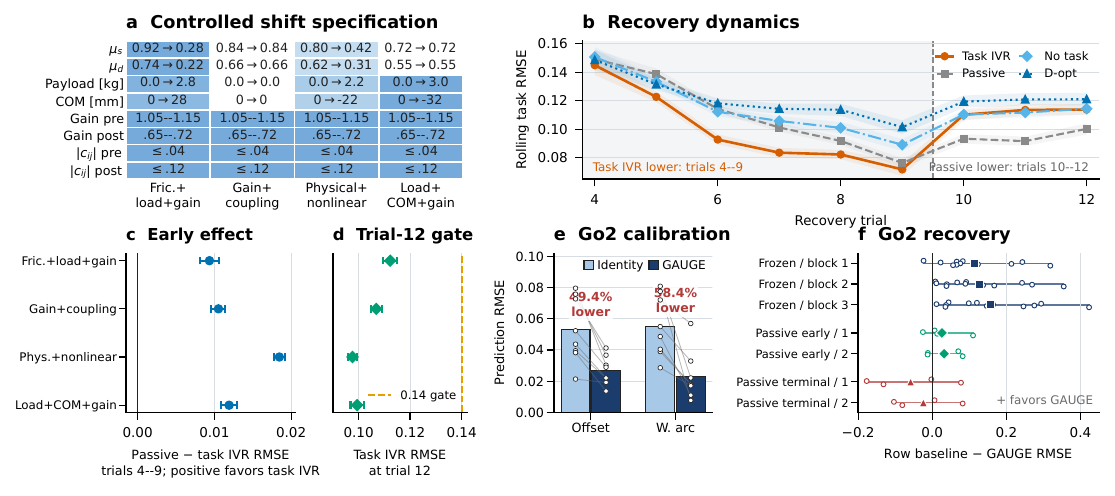}
  \caption{Calibration and recovery under interface shifts. (a) Shift parameters; blue denotes row-normalized change magnitude. (b) Recovery dynamics; shading marks trials 4--9. (c--d) Independent early-effect and trial-12 estimates. Simulation intervals are paired 95\% bootstrap intervals (30 seeds in b; 144 per shift in c--d). (e) Online calibration: bars show means; linked circles show matched commands. (f) Scheduled-hardware recovery: open circles show command contrasts, filled markers show block means, and spans show observed ranges. Positive contrasts favor \method; hardware effects are descriptive.}
  \label{fig:recovery}
\end{figure*}

Two route-specific Go2 deployments next test online response-model calibration.
In each deployment, six seed trials precede six commands selected by
task-weighted IVR, and eight fixed commands disjoint from calibration evaluate
the frozen posterior. Relative to the identity-map baseline, mean held-out
response-prediction RMSE falls from \RealOffsetRawRmse\ to
\RealOffsetGaugeRmse\ on offset slalom and from \RealArcRawRmse\ to
\RealArcGaugeRmse\ on weighted arc, reductions of \RealOffsetReduction\ and
\RealArcReduction\ (Fig.~\ref{fig:recovery}e). Both route-level
comparisons favor the fitted map, and all \RealCalibrationProbeCount\ matched
command-level observations have lower pointwise prediction error. Each
method--route cell was observed once, so the percentages are descriptive
route-level effects and commands are repeated measurements within a deployment.
The response model is refit independently for each route. These deployments
establish online feasibility and held-out response improvement. Paired selector
comparisons are isolated in the controlled studies below.

Controlled closed-loop runs test whether calibration improves realized motion
under physics. Four Isaac Lab contexts \cite{mittal2025isaaclab} use 20 paired
seeds, 12 calibration commands, and eight held-out commands with a common robust
constant-twist estimator. Raw-to-calibrated RMSE is
0.15199$\rightarrow$0.10635 for affine distortion,
0.15054$\rightarrow$0.10282 for dead-zone distortion,
0.10741$\rightarrow$0.09742 for low friction with load and center-of-mass
offset, and 0.09554$\rightarrow$0.07849 on rough terrain with load and offset.
Relative reductions span 9.30--31.70\%, and all
20 paired seeds improve in every context. These results support using the external-response model under the tested
response and terrain distortions. The next experiment evaluates acquisition
efficiency.

\subsection{Planner Conditioning Reduces Calibration Effort}

The primary experiment tests whether conditioning on the planner measure adds
sample efficiency beyond generic sequential design. It uses 20 paired seeds in
affine, dead-zone, and heteroscedastic synthetic families. All methods share a
six-command seed and a 160-trial horizon over
$[-1,1]\times[-.5,.5]\times[-1.5,1.5]$, with
$\|(v_x,v_y)\|_2\leq1$. Frozen draws vary gain, off-axis coupling, bias,
saturation, and noise. The declared task measure weights forward, left-turn,
and right-turn command components by $(.50,.25,.25)$. The stopping target is
RMSE $\leq0.04$ and integrated epistemic variance $\leq0.0015$. No-task IVR is
the direct ablation of planner conditioning. Random, Latin hypercube sampling
(LHS), Sobol, D-optimal, and a dense reference provide passive and
parameter-oriented controls. Acquisition penalties are zero. A hidden
400-command grid defines crossing and a disjoint 1,024-command grid audits it.
Paired Wilcoxon contrasts use Holm correction, and paired bootstrap intervals
use 4,000--5,000 resamples.

Task IVR reaches the joint accuracy--uncertainty criterion in 18.7 trials
(Fig.~\ref{fig:calibration}a). It requires fewer trials than the two closest
controls, D-optimal design and task-agnostic IVR, while fixed-budget held-out
RMSE preserves the same ordering (Fig.~\ref{fig:calibration}b). When evaluation
expands to the full authorized envelope, however, the ordering reverses
(Fig.~\ref{fig:calibration}c). Task conditioning therefore reallocates accuracy
toward the declared deployment support rather than improving the response model
uniformly.

\subsection{Task Conditioning Prioritizes Early Post-Shift Accuracy}

The shift study tests whether the original planner measure can direct
reacquisition after the response map changes. Four held-out shifts combine
friction, load, gain, coupling, nonlinear response, and center-of-mass effects
(Fig.~\ref{fig:recovery}a). The frozen deployment extension commissions
one residual signature at each of four labeled monitor commands
$\vect{c}_k=[v_{x,k},v_{y,k},\omega_{z,k}]^\top$, where $v_x$ and $v_y$ are in
$\mathrm{m\,s^{-1}}$ and $\omega_z$ is in $\mathrm{rad\,s^{-1}}$:
\begin{equation*}
\begin{aligned}
 \vect{c}_1&=(-.20,0,-.20)^\top, &
 \vect{c}_2&=(-.20,0,.20)^\top,\\
 \vect{c}_3&=(.12,-.10,-.15)^\top, &
 \vect{c}_4&=(.12,.10,.15)^\top.
\end{aligned}
\end{equation*}
It then cycles these commands over five post-shift monitor trials. It uses
$S=\operatorname{diag}(.14,.08,.18)$ and latches when $d_t>.30$ occurs in at
least two of the last four valid trials after a two-trial minimum dwell. A latch
retains the posterior mean, multiplies its covariance by eight, and opens a
12-command task-weighted recovery budget. Fixed-model, passive-update, and
\method\ arms share this detector schedule, 12 initial commands, and rolling
validation. Each shift pools two disjoint 72-seed blocks; the pre-specified
early endpoint averages recovery trials 4--9.

A matched selector ablation exposes how the ordering changes over the recovery
budget. Undefined trial-4 rolling windows receive the registered penalty of 0.25.
Across four shift contexts averaged within each of 30 paired seeds,
task IVR has the lowest mean rolling task RMSE throughout the pre-specified
trials 4--9. Passive updating crosses below it immediately after this window
and remains lower at trial 12; task-agnostic IVR and D-optimal design do not
match the early task-IVR trajectory (Fig.~\ref{fig:recovery}b). An independent,
larger-scale confirmation pools two disjoint 72-seed blocks per shift. Task IVR
reduces early-window RMSE relative to passive updating in all four shifts, with
positive paired 95\% intervals (Fig.~\ref{fig:recovery}c), while every terminal
interval upper endpoint remains below 0.116 and the declared 0.14 gate
(Fig.~\ref{fig:recovery}d). These results establish an early-window advantage for task-conditioned
recovery, with passive updating attaining lower error at trial 12 in the
selector ablation.

The residual monitor remains quiescent in stationary sequences, detects all
four shift families reliably, and returns task-facing error below the recovery
gate in nearly every run (Table~\ref{tab:shift_monitor}).

\begin{table}[t]
\caption{Isaac Lab shift-monitor operating characteristics.}
\label{tab:shift_monitor}
\centering
\footnotesize
\setlength{\tabcolsep}{2.5pt}
\begin{tabular}{@{}lccc@{}}
\hline
Outcome & Count & Exact 95\% interval & Worst $p_{95}$ trials / s \\
\hline
False alarm & 0/120 & $[0,.0303]$ & --- \\
Detection & 143--144/144 & $[.9619,.9998]$ & 3.9 / 10.1 \\
Recovery & 141--144/144 & $[.9403,.9957]$ & 9.0 / 46.8 \\
\hline
\end{tabular}
\par\vspace{1pt}
\parbox{\columnwidth}{\scriptsize Detection and recovery counts span four
144-seed shifts; intervals are the lowest exact 95\% intervals and delays the
worst $p_{95}$. False alarms cover 120 stationary sequences (12,000 trials).
Seconds denote monitor time for detection and the fixed
recovery-plus-validation schedule for recovery.}
\end{table}

\begin{figure*}[t]
  \centering
  \setlength{\tabcolsep}{0.5pt}
  \begin{tabular}{@{}cccccc@{}}
    \includegraphics[width=0.16\textwidth]{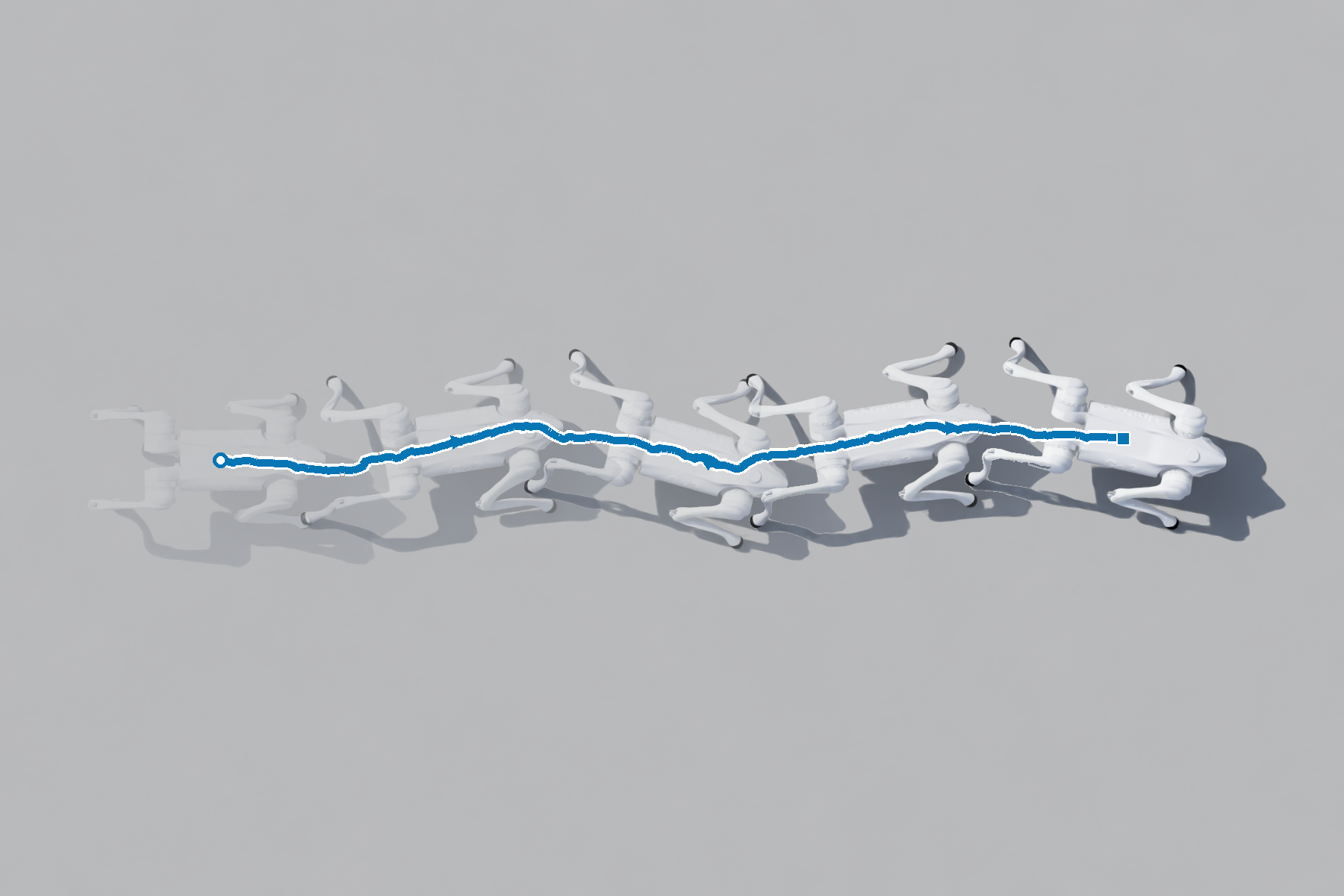} &
    \includegraphics[width=0.16\textwidth]{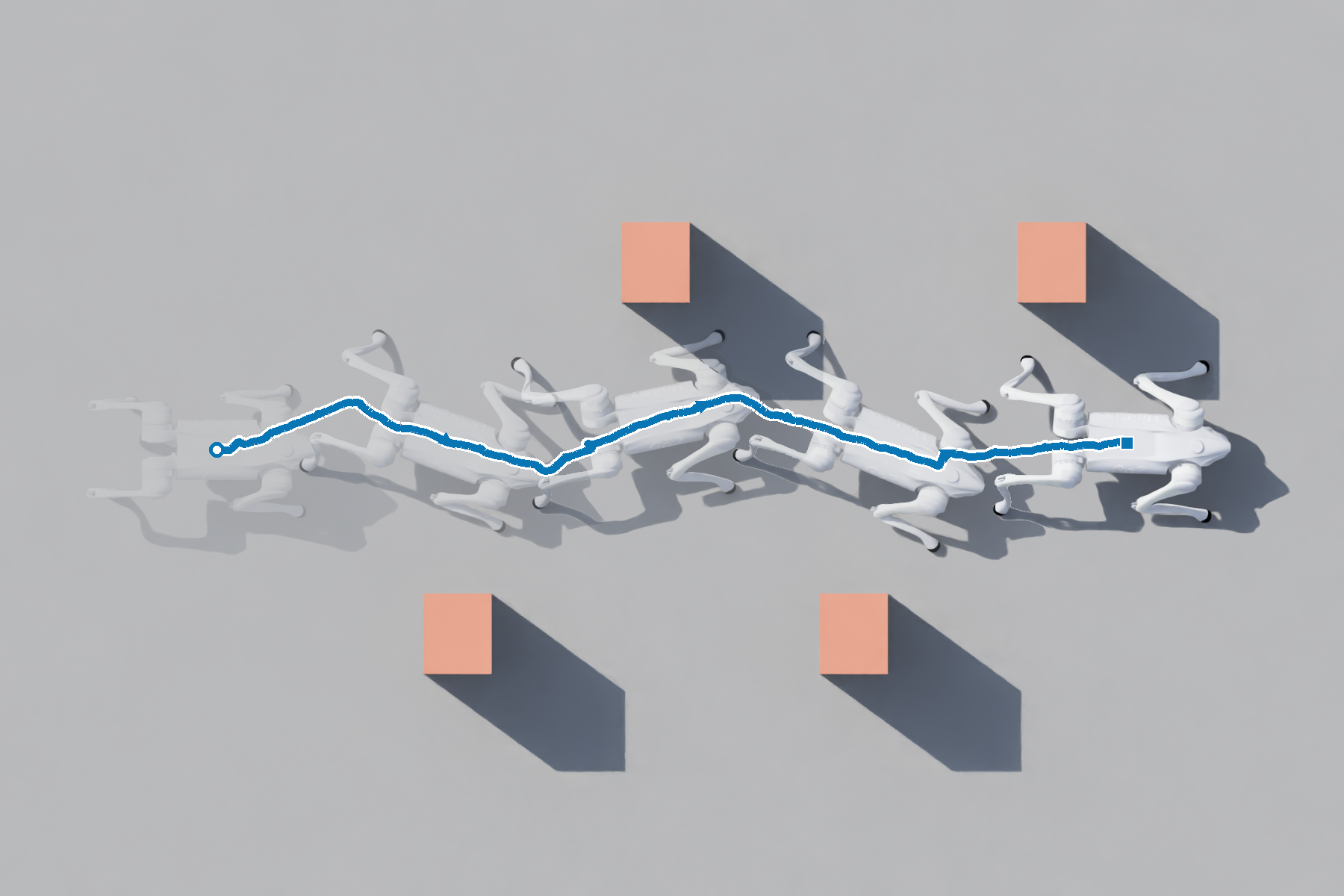} &
    \includegraphics[width=0.16\textwidth]{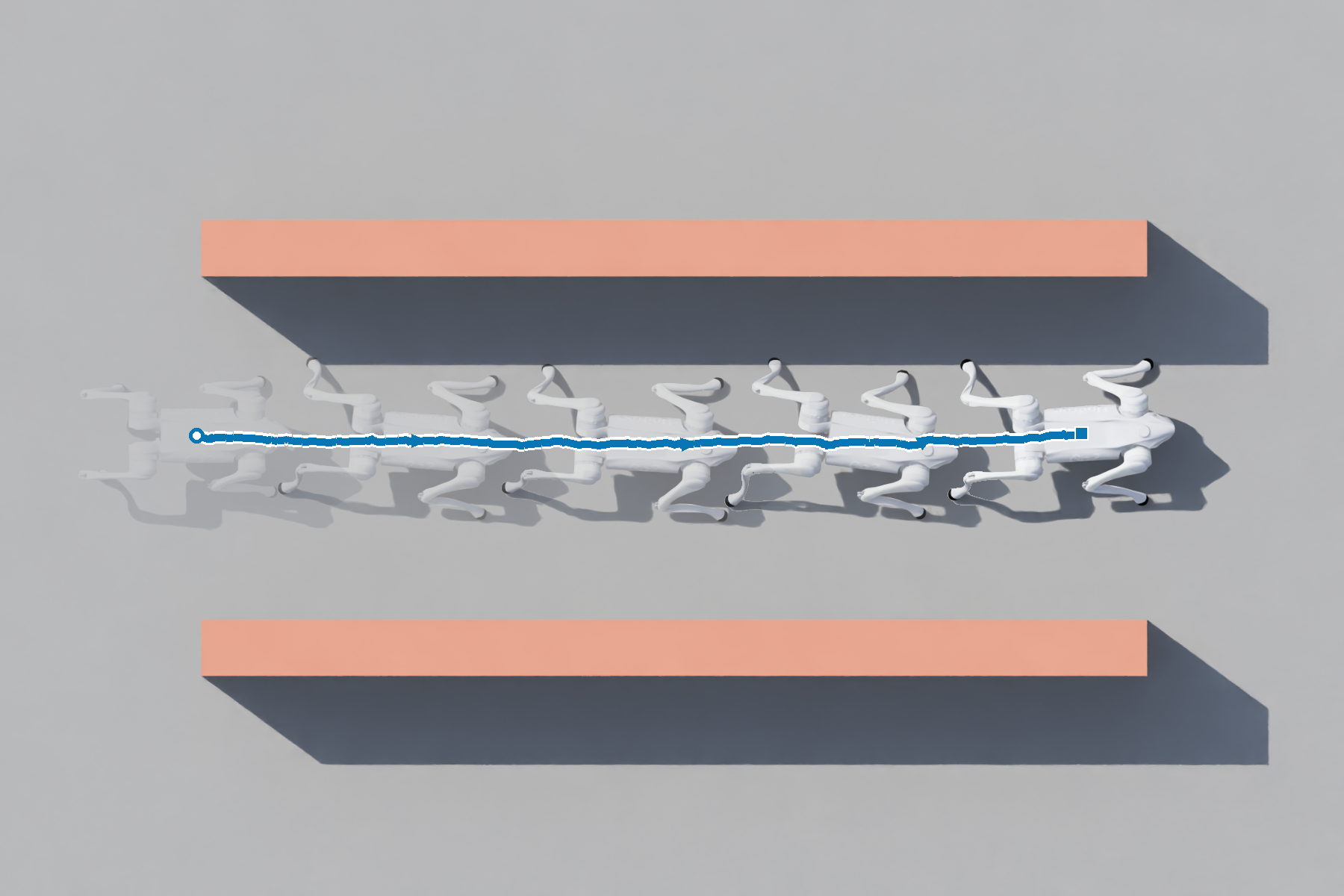} &
    \includegraphics[width=0.16\textwidth]{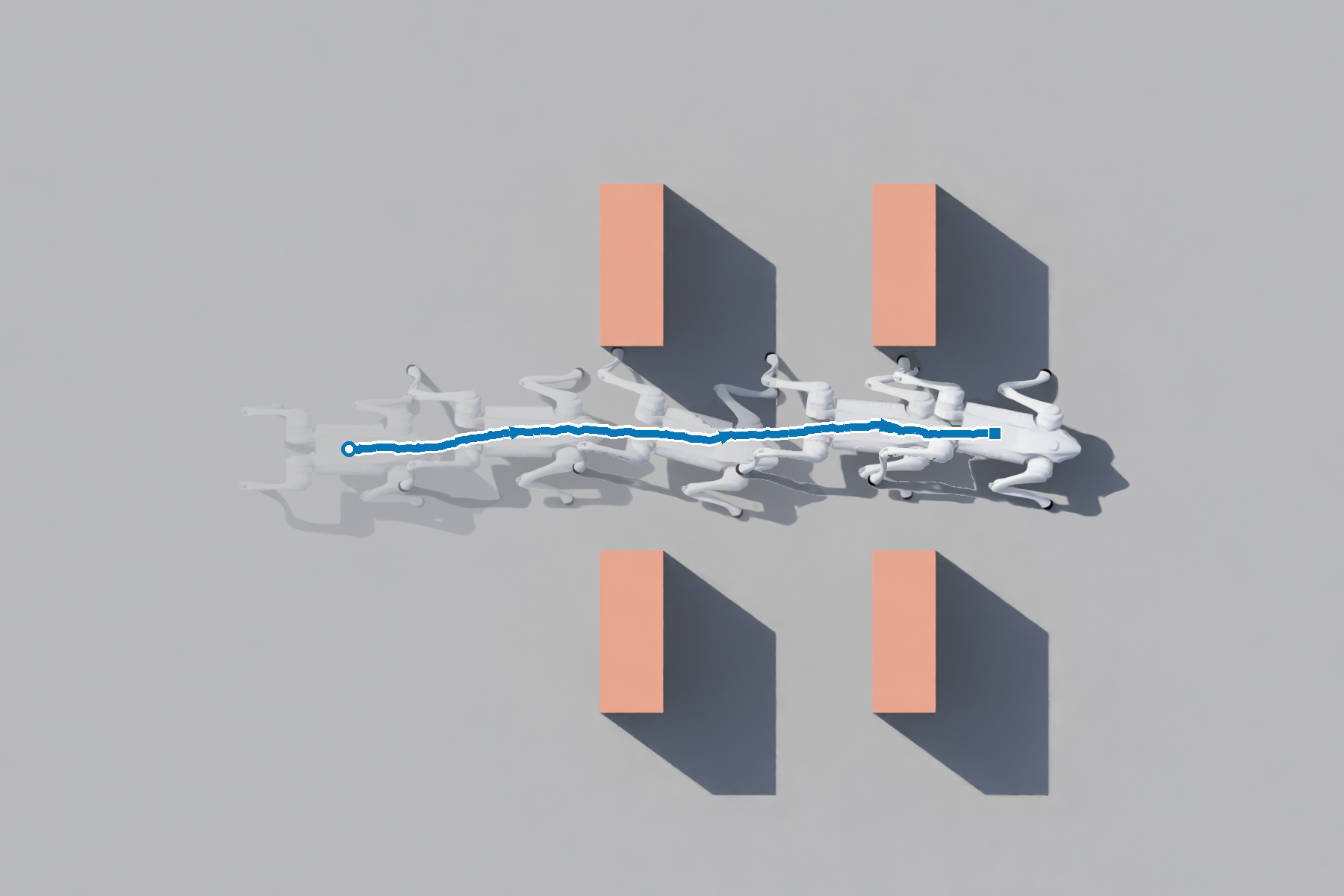} &
    \includegraphics[width=0.16\textwidth]{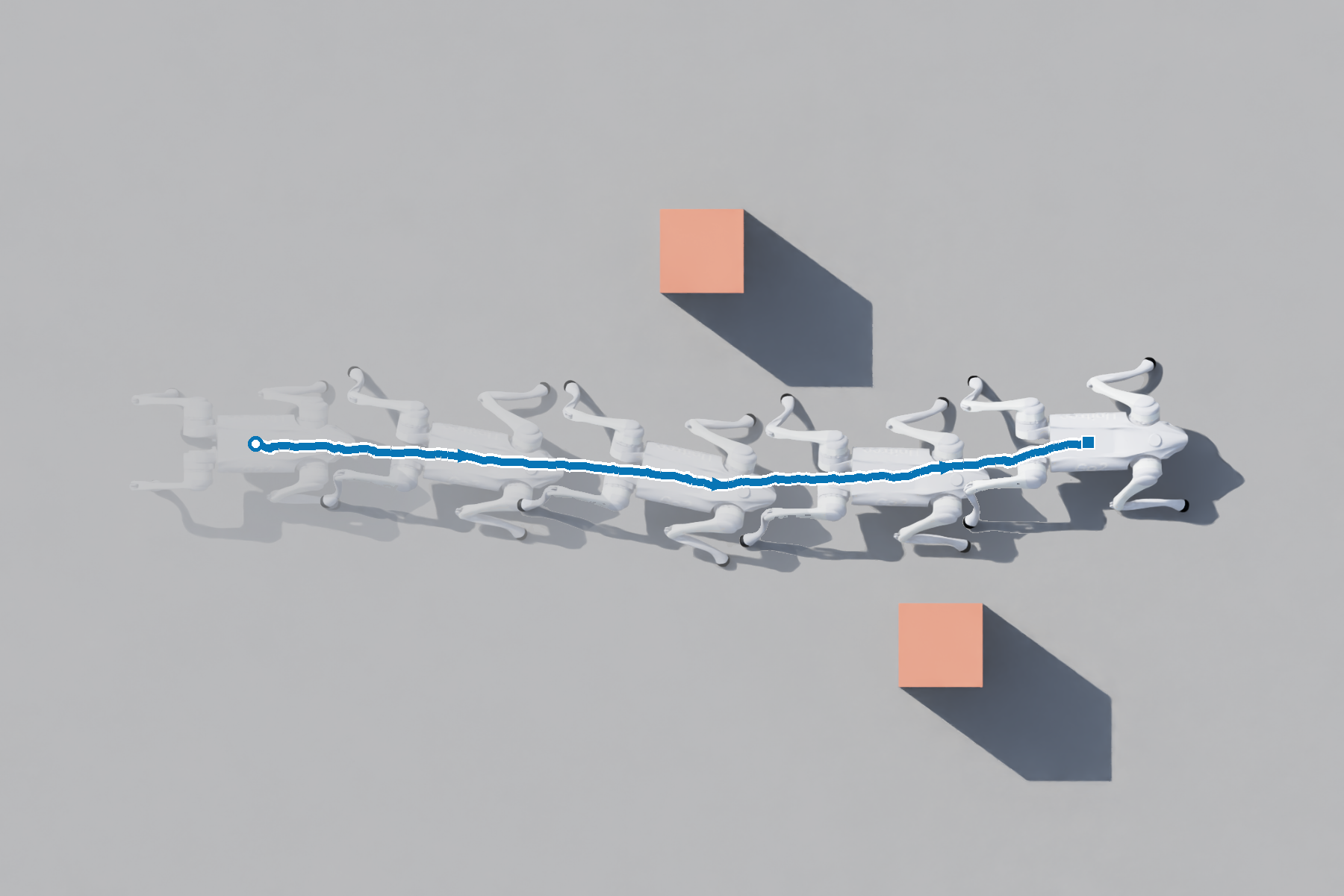} &
    \includegraphics[width=0.16\textwidth]{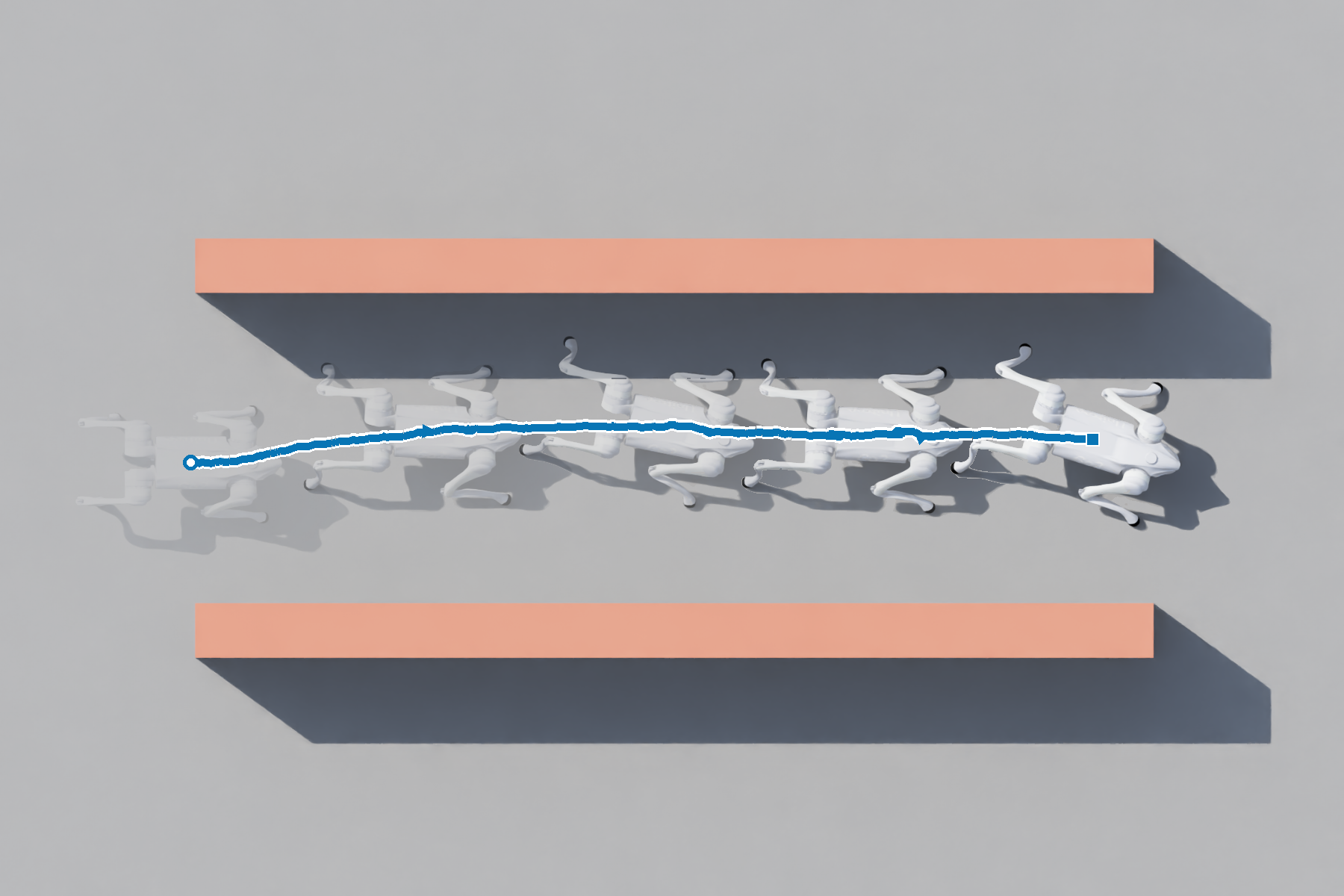} \\
    \scriptsize\sffamily S-bend & \scriptsize\sffamily Offset &
    \scriptsize\sffamily Narrow & \scriptsize\sffamily Chicane &
    \scriptsize\sffamily Weighted arc & \scriptsize\sffamily Extended
  \end{tabular}\\[-1pt]
  \includegraphics[width=0.97\textwidth]{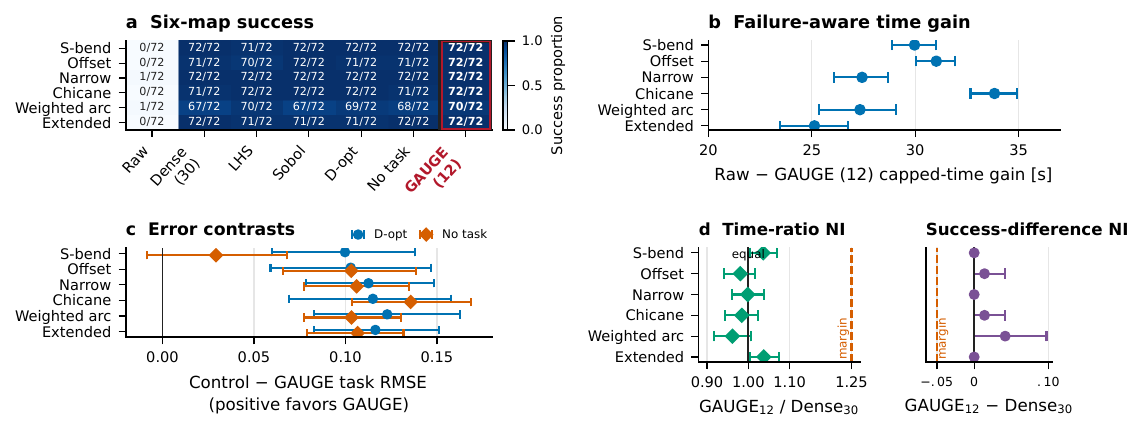}
  \caption{Navigation after calibration. Top: overhead replays on six held-out Isaac Sim maps. Blue curves show executed trajectories; opacity increases with time. Bottom: (a) success counts; (b) raw-minus-\method\ capped-time gain (60-s failure cap); (c) matched-budget RMSE contrasts; (d) noninferiority tests of \method\ (12 trials) versus dense calibration (30 trials), with time-ratio and success-difference margins of 1.25 and $-0.05$. Points and whiskers denote paired means and 95\% bootstrap intervals over 72 seeds per map.}
  \label{fig:navigation}
\end{figure*}

A scheduled gain-and-coupling perturbation on the Go2 changes the command gains
to $(0.82,0.90,1.15)$, with yaw-to-forward and forward-to-yaw couplings of 0.08
and 0.06, at a known intervention time. Frozen contrasts use 12 commands in each
of three blocks; passive early and terminal contrasts use the first and last
four commands in each of the two complete paired blocks. Following the
scheduled perturbation,
task-conditioned updating reduces process-averaged error relative to the frozen
map in all three blocks. In the two complete paired blocks, it also reduces the
pre-specified early endpoint relative to passive updating, whereas passive
updating attains lower terminal error (Fig.~\ref{fig:recovery}f). This physical
reversal is consistent with the controlled dynamics in
Fig.~\ref{fig:recovery}b and the intended recovery objective.

Operational checks verify that information ranking remains subordinate to
execution authorization. Across 60 acquisition traces, the hard filter rejects
300/300 hazardous proposals, rejects 0/20 authorized controls, and catches all
160 injected state faults. The stopping rule makes no premature decision on 60
trajectories. Median and p95 overshoot are two trials. The 50~Hz physics monitor
has a maximum 20~ms abort interval. These outcomes verify authorization and
stopping under replay and simulation.

\subsection{Task-Conditioned Accuracy Transfers to Navigation}

Navigation tests whether task-facing response accuracy changes planner outcomes
without dense global calibration. The paired study crosses six disjoint maps,
seven calibration methods, and 72 seeds per map, for 3,024 episodes
(Fig.~\ref{fig:navigation}). All arms share the controller, 50~Hz interlock,
10~Hz waypoint follower, geometry, and randomization. \method\ and matched
controls use 12 calibration commands, while the dense reference uses 30, followed by
eight common validation commands. Task IVR ranks a 512-command design pool under
the planner-facing measure. Routes span 2.85--3.60~m, with a 0.79~m minimum
opening and 0.20~m collision radius. Success requires reaching the 0.25~m goal
ball without collision or simulator termination by 60~s, and every failure
retains the 60-s cap. The declared noninferiority margins are a 1.25 capped-time
ratio and a $-0.05$ success difference.

At the same 12-trial budget, task IVR reduces validation RMSE over D-optimal by
9.98--12.28\% on every map. Its advantage over no-task IVR is positive on five
maps and unresolved on S-bend (Fig.~\ref{fig:navigation}c). The corresponding
\method\ policy succeeds in 72/72 episodes on five maps and 70/72 on weighted
arc, whereas raw commands succeed in 0/72 or 1/72
(Fig.~\ref{fig:navigation}a). Every \method\ collision count is zero. The 95\%
lower endpoints of the paired raw-minus-\method\ capped-time gain are
23.45--32.68~s (Fig.~\ref{fig:navigation}b).

The worst \method/dense capped-time ratio upper bound is 1.0736, below the 1.25
margin, and every paired success-difference lower bound exceeds $-0.05$
(Fig.~\ref{fig:navigation}d). The two weighted-arc failures are timeouts. Acquisition takes 31.2~s for \method\ and 78.0~s
for dense calibration, or 52.0 and 98.8~s after common validation. Across its
432 episodes, \method\ records 10 authorization interventions and no simulator
termination.

\begin{figure}[t]
  \centering
  \setlength{\tabcolsep}{0pt}
  \begin{tabular}{@{}c@{}}
    \includegraphics[width=\columnwidth]{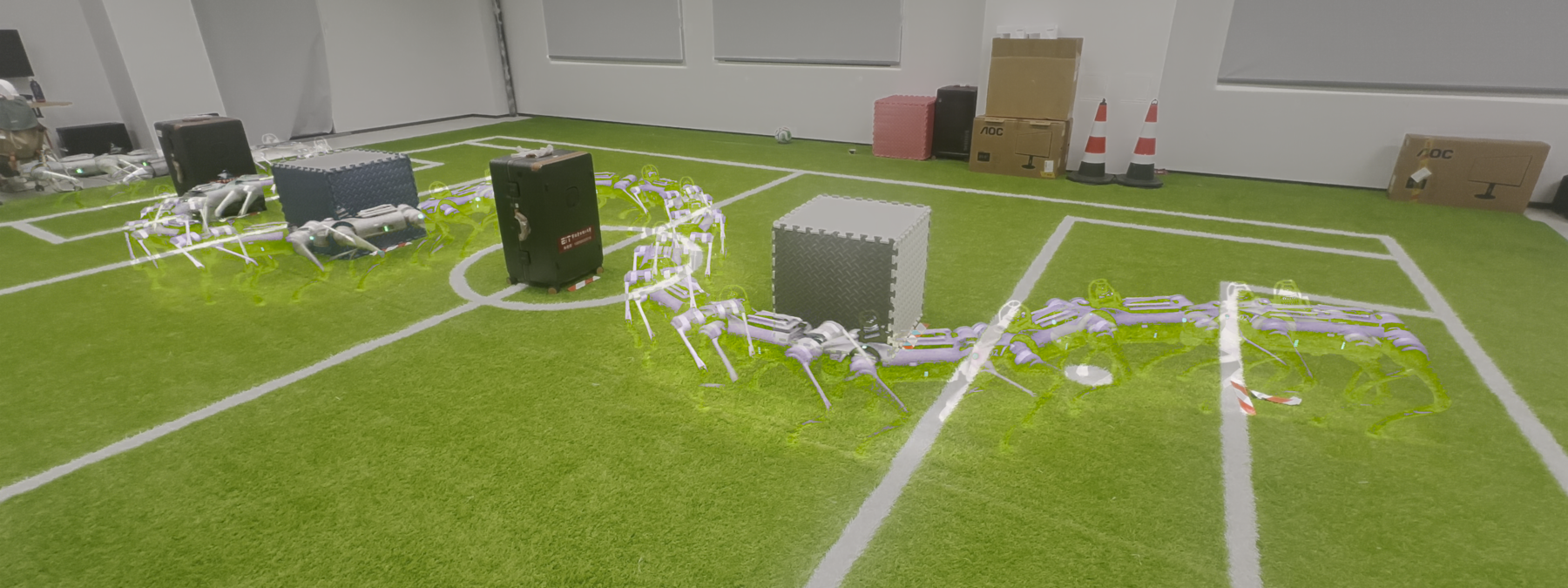} \\[-1pt]
    \footnotesize\textbf{(a)} Go2: offset slalom \\[3pt]
    \includegraphics[width=\columnwidth]{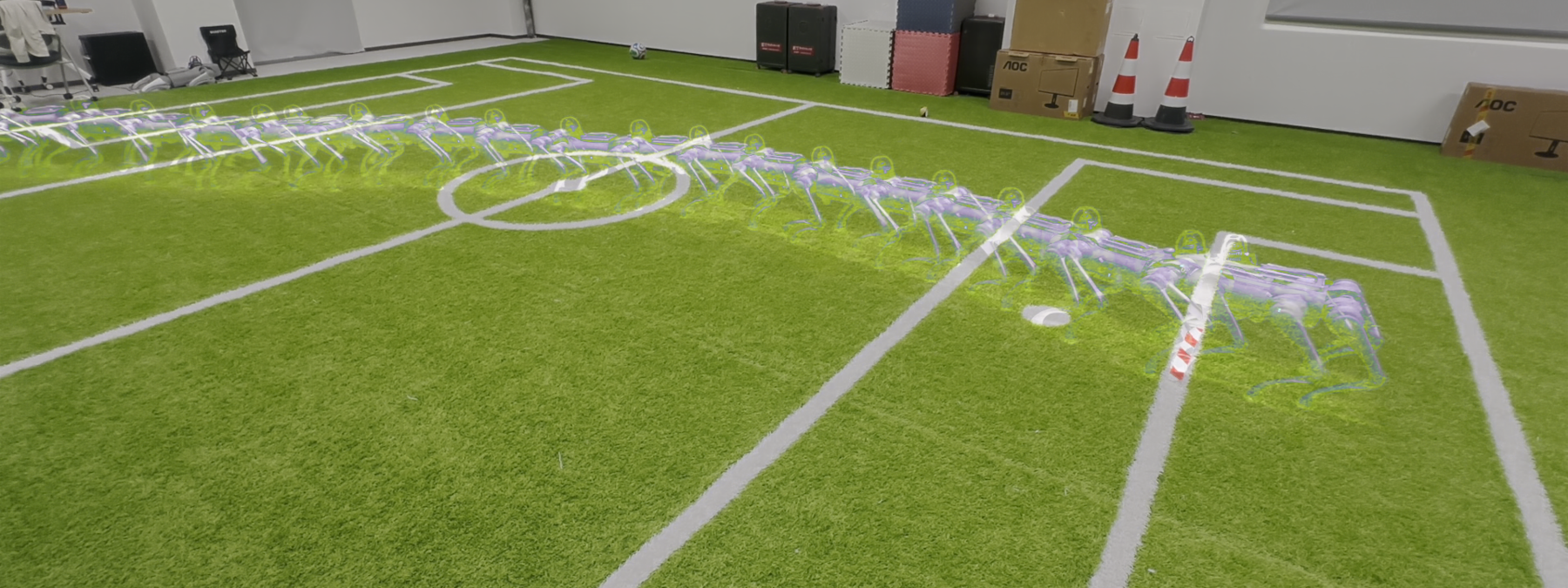} \\[-1pt]
    \footnotesize\textbf{(b)} Go2: weighted arc \\[3pt]
    \includegraphics[width=\columnwidth]{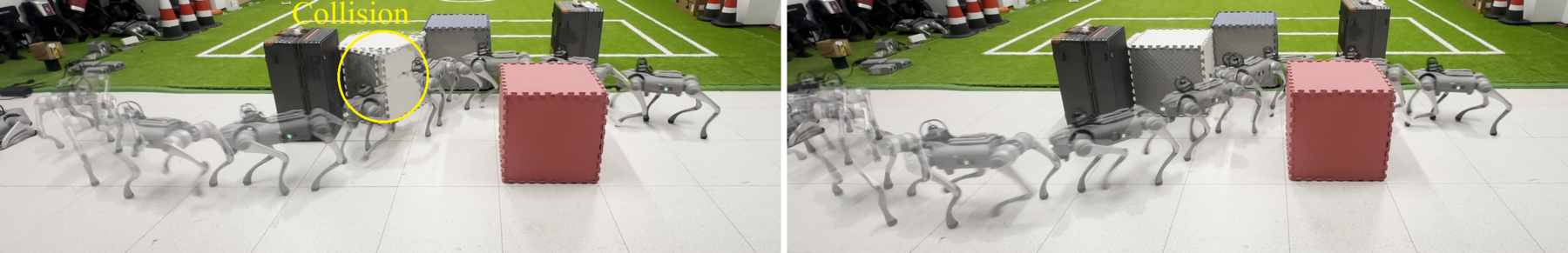} \\[-1pt]
    \footnotesize\textbf{(c)} DRL-DCLP: direct / \method
  \end{tabular}
  \caption{Physical Go2 navigation records. (a--b) Long-exposure views of the
  route-specific \method\ deployments on offset slalom and weighted arc.
  (c) A separate fixed-DRL-DCLP endurance-gait comparison in the same obstacle
  scene: the direct-interface rollout (left) contacts an obstacle, whereas the
  \method-calibrated rollout (right) clears it. These physical panels are
  qualitative route-level records.}
  \label{fig:physical_navigation}
\end{figure}

The physical navigation sweep uses a fixed DRL-DCLP local planner
\cite{zhang2025drldclp} and the same planner-to-Sport-Mode command chain for all
conditions. Eight interface conditions are executed once on each of two routes,
yielding \RealNavigationSuccess\ successful episodes, no logged collision, and
valid localization, scan, command-delivery, and start-state quality checks. The
two \method\ episodes follow their route-specific 12-trial online fits and are
visualized in Fig.~\ref{fig:physical_navigation}a--b. Separately, the fixed-planner
demonstration in Fig.~\ref{fig:physical_navigation}c uses the Go2 endurance gait in the
same obstacle scene: the observed direct-interface rollout contacts an
obstacle, whereas the \method-calibrated rollout clears it. This qualitative
comparison complements the formal 16-run deployment sweep. With one episode
per cell and saturated task success, the formal runs establish closed-loop
deployability. The 72-seed Isaac study provides the comparative navigation
effect.

\section{Discussion and Limitations}

Planner-conditioned calibration uses fewer trials to meet task-facing accuracy
and uncertainty targets in the tested response families. The broad-uniform
control reverses the held-out-error ordering, linking this gain to the match
between acquisition and deployment command distributions.

The same principle explains post-shift recovery. Planner conditioning restores
accuracy first where the planner acts, whereas passive updating can attain lower
error at the end of the recovery budget. The controlled selector ablation
isolates this early advantage, and the scheduled Go2 perturbation exhibits the
same early-versus-terminal trade-off in the two complete paired blocks. Recovery
therefore reuses the same task-scoped calibration objective after a shift.

\method\ assumes a steady command--response relation for a fixed gait and
controller. Strong history dependence, rapid transients, or controller switching
require additional response state or separate models, and a changed planner
distribution requires a new task measure. The physical experiments establish
online calibration and scheduled recovery; autonomous residual-triggered
recovery remains to be quantified on hardware.

\section{Conclusion}

Calibration of an opaque command interface should follow the command
distribution induced by its downstream planner. \method\ formalizes this
principle as task-facing posterior uncertainty and selects each authorized trial
for its expected reduction of that quantity. In controlled response families,
planner conditioning reaches the joint accuracy--uncertainty target with fewer
trials than passive, parameter-oriented, and task-agnostic acquisition. The
broad-uniform control establishes the support-dependent boundary of that gain.
On the Go2, route-specific 12-trial fits lower held-out response-prediction error,
and scheduled task-weighted updating lowers process-averaged post-shift error
relative to a frozen map in all three physical blocks. After shifts, task-conditioned updating lowers early task-facing error
relative to passive updating, while held-out Isaac navigation meets the declared noninferiority margins
against dense calibration across six maps. Within its declared task support,
\method\ makes the external velocity interface a calibrated control variable
for the planner.

\bibliographystyle{IEEEtran}
\bibliography{references}

\begin{thebibliography}{10}
\providecommand{\url}[1]{#1}
\csname url@rmstyle\endcsname
\providecommand{\newblock}{\relax}
\providecommand{\bibinfo}[2]{#2}
\providecommand\BIBentrySTDinterwordspacing{\spaceskip=0pt\relax}
\providecommand\BIBentryALTinterwordstretchfactor{4}
\providecommand\BIBentryALTinterwordspacing{\spaceskip=\fontdimen2\font plus
\BIBentryALTinterwordstretchfactor\fontdimen3\font minus
  \fontdimen4\font\relax}
\providecommand\BIBforeignlanguage[2]{{%
\expandafter\ifx\csname l@#1\endcsname\relax
\typeout{** WARNING: IEEEtran.bst: No hyphenation pattern has been}%
\typeout{** loaded for the language `#1'. Using the pattern for}%
\typeout{** the default language instead.}%
\else
\language=\csname l@#1\endcsname
\fi
#2}}

\bibitem{hwangbo2019agile}
J.~Hwangbo, J.~Lee, A.~Dosovitskiy, C.~D. Bellicoso, V.~Tsounis, V.~Koltun, and
  M.~Hutter, ``Learning agile and dynamic motor skills for legged robots,''
  \emph{Science Robotics}, vol.~4, no.~26, p. eaau5872, 2019.

\bibitem{taouil2023blackbox}
I.~Taouil, G.~Turrisi, D.~Schleich, V.~Barasuol, C.~Semini, and S.~Behnke,
  ``Quadrupedal footstep planning using learned motion models of a black-box
  controller,'' in \emph{Proc. IEEE/RSJ Int. Conf. Intelligent Robots and
  Systems (IROS)}, 2023, pp. 800--806.

\bibitem{lee2020terrain}
J.~Lee, J.~Hwangbo, L.~Wellhausen, V.~Koltun, and M.~Hutter, ``Learning
  quadrupedal locomotion over challenging terrain,'' \emph{Science Robotics},
  vol.~5, no.~47, p. eabc5986, 2020.

\bibitem{rudin2022minutes}
N.~Rudin, D.~Hoeller, P.~Reist, and M.~Hutter, ``Learning to walk in minutes
  using massively parallel deep reinforcement learning,'' in \emph{Proc. Conf.
  Robot Learning}, ser. Proceedings of Machine Learning Research, vol. 164,
  2022, pp. 91--100.

\bibitem{curtis2025flow}
A.~Curtis, E.~Li, M.~Noseworthy, N.~Gothoskar, S.~Chitta, H.~Li, L.~P.
  Kaelbling, and N.~E. Carey, ``Flow-based domain randomization for learning
  and sequencing robotic skills,'' in \emph{Proc. Int. Conf. Machine Learning},
  ser. Proceedings of Machine Learning Research, vol. 267, 2025, pp.
  11\,692--11\,709.

\bibitem{kumar2021rma}
A.~Kumar, Z.~Fu, D.~Pathak, and J.~Malik, ``{RMA}: Rapid motor adaptation for
  legged robots,'' in \emph{Robotics: Science and Systems XVII}, 2021.

\bibitem{fey2024adapt}
N.~P. Fey, H.~Li, N.~Adrian, P.~M. Wensing, and M.~Lemmon, ``A learning-based
  framework to adapt legged robots on-the-fly to unexpected disturbances,'' in
  \emph{Proc. Learning for Dynamics and Control Conf.}, ser. Proceedings of
  Machine Learning Research, vol. 242, 2024, pp. 1161--1173.

\bibitem{li2026embodiment}
D.~Li, B.~Ai, N.~Bohlinger, J.~Peters, H.~Su, and H.~I. Christensen, ``Rapid
  embodiment adaptation for quadrupedal locomotion,'' \emph{arXiv preprint
  arXiv:2608.01506}, 2026.

\bibitem{sun2021unknown}
Y.~Sun, W.~L. Ubellacker, W.-L. Ma, X.~Zhang, C.~Wang, N.~Csomay-Shanklin,
  M.~Tomizuka, K.~Sreenath, and A.~D. Ames, ``Online learning of unknown
  dynamics for model-based controllers in legged locomotion,'' \emph{IEEE
  Robotics and Automation Letters}, vol.~6, no.~4, pp. 8442--8449, 2021.

\bibitem{widmer2023safe}
\BIBentryALTinterwordspacing
D.~Widmer, D.~Kang, B.~Sukhija, J.~H\"{u}botter, A.~Krause, and S.~Coros,
  ``Tuning legged locomotion controllers via safe bayesian optimization,'' in
  \emph{Proceedings of the 7th Conference on Robot Learning}, ser. Proceedings
  of Machine Learning Research, vol. 229, 2023, pp. 2444--2464. [Online].
  Available: \url{https://proceedings.mlr.press/v229/widmer23a.html}
\BIBentrySTDinterwordspacing

\bibitem{haack2025adaptive}
J.~Haack, F.~Stark, S.~Vyas, F.~Kirchner, and S.~Kumar, ``Adaptive model-based
  control of quadrupeds via online system identification using kalman filter,''
  in \emph{Proc. IEEE/RSJ Int. Conf. Intelligent Robots and Systems (IROS)},
  2025, pp. 5039--5044.

\bibitem{li2022vio}
H.~Li and J.~St\"uckler, ``Visual-inertial odometry with online calibration of
  velocity-control based kinematic motion models,'' \emph{IEEE Robotics and
  Automation Letters}, vol.~7, no.~3, pp. 6415--6422, 2022.

\bibitem{hollerbach1996calibration}
J.~M. Hollerbach and C.~W. Wampler, ``The calibration index and taxonomy for
  robot kinematic calibration methods,'' \emph{The International Journal of
  Robotics Research}, vol.~15, no.~6, pp. 573--591, 1996.

\bibitem{calafiore2001dynamic}
G.~Calafiore, M.~Indri, and B.~Bona, ``Robot dynamic calibration: Optimal
  excitation trajectories and experimental parameter estimation,''
  \emph{Journal of Robotic Systems}, vol.~18, no.~2, pp. 55--68, 2001.

\bibitem{sun2008observability}
Y.~Sun and J.~M. Hollerbach, ``Observability index selection for robot
  calibration,'' in \emph{Proc. IEEE Int. Conf. Robotics and Automation
  (ICRA)}, 2008, pp. 831--836.

\bibitem{sun2008active}
------, ``Active robot calibration algorithm,'' in \emph{Proc. IEEE Int. Conf.
  Robotics and Automation (ICRA)}, 2008, pp. 1276--1281.

\bibitem{carrillo2013task}
H.~Carrillo, O.~Birbach, H.~T\"aubig, B.~B\"auml, U.~Frese, and J.~A.
  Castellanos, ``On task-oriented criteria for configurations selection in
  robot calibration,'' in \emph{Proc. IEEE Int. Conf. Robotics and Automation
  (ICRA)}, 2013, pp. 3653--3659.

\bibitem{attia2018goal}
A.~Attia, A.~Alexanderian, and A.~K. Saibaba, ``Goal-oriented optimal design of
  experiments for large-scale bayesian linear inverse problems,'' \emph{Inverse
  Problems}, vol.~34, no.~9, p. 095009, 2018.

\bibitem{sobanbabu2025spiactive}
\BIBentryALTinterwordspacing
N.~Sobanbabu, G.~He, T.~He, Y.~Yang, and G.~Shi, ``Sampling-based system
  identification with active exploration for legged {Sim2Real} learning,'' in
  \emph{Proceedings of the 9th Conference on Robot Learning}, ser. Proceedings
  of Machine Learning Research, vol. 305, 2025, pp. 578--598. [Online].
  Available: \url{https://proceedings.mlr.press/v305/sobanbabu25a.html}
\BIBentrySTDinterwordspacing

\bibitem{krause2008sensor}
A.~Krause, A.~Singh, and C.~Guestrin, ``Near-optimal sensor placements in
  gaussian processes: Theory, efficient algorithms and empirical studies,''
  \emph{Journal of Machine Learning Research}, vol.~9, pp. 235--284, 2008.

\bibitem{xu2021fastlio}
W.~Xu and F.~Zhang, ``{FAST-LIO}: A fast, robust {LiDAR}-inertial odometry
  package by tightly-coupled iterated {Kalman} filter,'' \emph{IEEE Robotics
  and Automation Letters}, vol.~6, no.~2, pp. 3317--3324, 2021.

\bibitem{mittal2025isaaclab}
M.~Mittal \emph{et~al.}, ``Isaac lab: A gpu-accelerated simulation framework
  for multi-modal robot learning,'' \emph{arXiv preprint arXiv:2511.04831},
  2025.

\bibitem{zhang2025drldclp}
W.~Zhang, S.~Wang, M.~Tan, Z.~Yang, X.~Wang, and X.~Shen, ``{DRL-DCLP}: A deep
  reinforcement learning-based dimension-configurable local planner for robot
  navigation,'' \emph{IEEE Robotics and Automation Letters}, vol.~10, no.~4,
  pp. 3636--3643, 2025.

\end{thebibliography}

\end{document}